\documentclass[times,twocolumn,final]{elsarticle}
\usepackage{framed,multirow}
\usepackage{algorithm}
\usepackage{algpseudocode}
\usepackage{amssymb}
\usepackage{latexsym}
\usepackage{url}
\usepackage{amsfonts} 
\usepackage{amsmath}
\usepackage{amsthm}
\usepackage{pifont}
\usepackage{natbib}
\usepackage{geometry}
\usepackage{graphicx}
\usepackage{color} 
 
\newcommand{\subfigimg}[3][,]{%
  \setbox1=\hbox{\includegraphics[#1]{#3}}
  \leavevmode\rlap{\usebox1}
  \rlap{\hspace*{12pt}\raisebox{\dimexpr\ht1-0\baselineskip}{#2}}
  \phantom{\usebox1}
}
\RequirePackage{lineno}
\usepackage{subfig}

\newenvironment{Table}
  {\par\bigskip\noindent\minipage{\columnwidth}\centering}
  {\endminipage\par\bigskip}
	
\usepackage{tikz}
\usetikzlibrary{shapes.geometric, arrows}
\tikzstyle{node} = [rectangle, rounded corners, minimum width=0.5cm, minimum height=0.5cm,text centered, draw=black, fill=white!30]
\tikzstyle{arrow} = [thick,->,>=stealth]

\newcommand{\Linefor}[2]{%
    \State \algorithmicfor\ {#1}\ \algorithmicdo\ {#2} \algorithmicend\ \algorithmicfor%
}

\newcommand{\rto}{\leftarrow}
\providecommand{\mb}[1]{\boldsymbol{#1}}

\begin{document}

\begin{frontmatter}
\title{Manifold Matching using Shortest-Path Distance and Joint Neighborhood Selection}
\author[1,2]{Cencheng Shen}
\ead{cshen6@jhu.edu}
\author[1,3]{Joshua T. Vogelstein}
\ead{jovo@jhu.edu}
\author[1,4]{Carey E. Priebe\corref{cor1}}
\ead{cep@jhu.edu}

\address[1]{Center for Imaging Science, Johns Hopkins University}
\address[2]{Department of Statistics, Temple University}
\address[3]{Department of Biomedical Engineering and Institute for Computational Medicine, Johns Hopkins University}
\address[4]{Department of Applied Mathematics and Statistics, Johns Hopkins University}
\cortext[cor1]{Corresponding author}

\begin{abstract}
Matching datasets of multiple modalities has become an important task in data analysis. Existing methods often rely on the embedding and transformation of each single modality without utilizing any correspondence information, which often results in sub-optimal matching performance. In this paper, we propose a nonlinear manifold matching algorithm using shortest-path distance and joint neighborhood selection. Specifically, a joint nearest-neighbor graph is built for all modalities. Then the shortest-path distance within each modality is calculated from the joint neighborhood graph, followed by embedding into and matching in a common low-dimensional Euclidean space. Compared to existing algorithms, our approach exhibits superior performance for matching disparate datasets of multiple modalities. 
\end{abstract}
\end{frontmatter}

\section{Introduction}
The abundance of data in the modern age has made it crucial to effectively deal with large amounts of high-dimensional data. For the purpose of data analysis, it is imperative to apply dimension reduction to embed data into a low-dimensional space for subsequent analysis. Traditional linear embedding techniques have solid theoretical foundations and are widely used, e.g., principal component analysis (PCA) \citep{JolliffePCABook, BishopTipping1999} and multidimensional scaling (MDS) \citep{TorgersonBook, BorgBook, CoxBook} for datasets of a single modality, and canonical correlation analysis (CCA) \citep{Hotelling1936, BachJordan2005} for datasets of multiple modalities. 

However, real datasets often exhibit nonlinear geometry, discovering which can be advantageous for subsequent inference. Many manifold learning algorithms have been proposed to learn the intrinsic low-dimensional structure of nonlinear datasets, including Isomap \citep{TenenbaumSilvaLangford2000, SilvaTenenbaum2003}, locally linear embedding (LLE) \citep{SaulRoweis2000, RoweisSaul2003}, Hessian LLE \citep{DonohoGrimes2003}, Laplacian eigenmaps \citep{BelkinNiyogi2003, HeEtAl2005}, local tangent space alignment (LTSA) \citep{ZhangZha2004, ZhangWangZha2012}, among many others. Most of them start with the assumption that the data are locally linear, explore the local geometry via the nearest-neighbor graph of the sample data, transform the data using the neighborhood graph, and eventually learn the low-dimensional manifold by optimizing some objective function. These nonlinear embedding algorithms usually serve as a preliminary feature extraction step that enables subsequent inference. They have been used successfully in object recognition and image processing. 

In this paper, we consider the manifold matching task for datasets of multiple modalities, which is traditionally modeled by multiple dependent random variables. Conventional methods for identifying the relationship among multiple random variables are still very popular in theory and practice, such as canonical correlation \citep{Hotelling1936, Kettenring1971, Hardoon2004} and Procrustes transformation \citep{Sibson1978, Sibson1979, GoldbergRitov2009, GowerProcrustesBook}. However, it has become a much more challenging task to match real datasets of multiple modalities from disparate sources due to their complex dependency structures, such as the same document in different languages, an image and its descriptions, or networks of the same actors on different social websites. 

There have been many recent endeavors regarding data fusion and manifold matching \citep{LafonKellerCoifman2006, WangMahadevan2008, WangMahadevan2012, SharmaKumar2012, PriebeMarchette2012, SunPriebeTang2013, ShenSunTangPriebe2014}. Similar to dimension reduction for datasets of a single modality, manifold matching can serve as a feature extraction step to explore datasets of multiple modalities, and has also been shown to help subsequent inference in object recognition \citep{KimKittlerCipolla2007}, information retrieval \citep{SunPriebe2012}, and transfer learning \citep{PanYang2010}. Furthermore, the matching task is important on its own and has been applied to explore multiple graphs and networks \citep{LyzinskiFishkindPriebe2014, JoshuaEtAl2015, LyzinskiFishkindPriebe2016}. One such application is seeded graph matching, where two large networks are collected but only a percentage of training vertices have known correspondence. Then the remaining vertices need to be properly matched to uncover potential correspondence and benefit later inference.

Due to the success of nonlinear embedding algorithms for datasets of a single modality, it is often perceived that these algorithms can be directly combined into the matching framework to improve the matching performance when one or more modalities are nonlinear. A na\"ive procedure is to pick one nonlinear algorithm, apply it to each modality separately, and match the embedded modalities. But such a simplistic procedure does not always guarantee a good matching performance, since many nonlinear embedding algorithms only preserve the local geometry up to some affine transformation \citep{GoldbergRitov2008}. Furthermore, using nonlinear transformations separately can even deteriorate the matching performance when compared to using simple linear transformations, as shown in our numerical simulations.

To tackle the problem, we propose a manifold matching algorithm using shortest-path distance and joint neighborhood selection. By utilizing a robust distance measure that approximates the geodesic distance, and effectively combining the correspondence information into the embedding step, the proposed algorithm can significantly improve the matching quality from disparate data sources, compared to directly take linear or nonlinear embeddings for matching. All code and data are made publicly available \footnote{\url{https://github.com/cshen6/MMSJ}}.

\section{Manifold Matching}
\label{main}

In this section, the matching framework and evaluation criteria are first introduced. Next we present the main algorithm, followed by relevant implementation details. Additional discussions are offered on issues that can affect the matching performance.

\subsection{The Matching Framework}
\label{bg}
Suppose $n$ objects are measured under two different sources. Then $X_{l}=\{x_{il}\} \in \Xi_{l}$ for $l=1, 2$ are the actual datasets that are observed / collected, with $x_{i1} \sim x_{i2}$ for each $i$ ($\sim$ means the two observations are matched in the context). Thus $X_{1}$ and $X_{2}$ are the two different views / modalities of the same underlying data. This setting is extendable to datasets of more than two modalities, but for ease of presentation we focus mainly on the matching task of two modalities.

$\Xi_{1}$ and $\Xi_{2}$ are potentially very different from each other, such as a flat manifold and its nonlinear transformation, an image and its description, or texts under different languages. A typical example is the social network, where many users have accounts on Youtube, Facebook, Twitter, etc. People sometimes post different contents and connect with different groups on each network site, such that data analysis of better quality is only possible when multiple accounts of the same person are combined. Some users already linked their accounts from different places, or unique user information are filled (like actual name, occupation), certain accounts can be automatically matched, providing a set of matched training data; but all the other accounts need to be matched by machine (as manual match is too expensive for millions of accounts), presenting a set of testing data from each website.

We assume $x_{il} \in \Xi_{l}$ is endowed with a distance measure $\Delta_{l}$ such that $\Delta_{l}(i,j)=dist(x_{il},x_{jl})$. To match multiple modalities, we find two mappings $\rho_{l}: \Xi_{l} \rightarrow \mathbb{R}^{d}, l=1,2$ such that the mapped data $\hat{X}_{l}=\{ \rho_{l}(x_{il}) \}$ are matched into a common low-dimensional Euclidean space $\mathbb{R}^{d}$. A simple example of $\rho_{l}$ is MDS (e.g., classical MDS first doubly centers the distance matrices, followed by eigen-decomposition and keeping the top $d$ eigenvalues and eigenvectors to yield the embedding) followed by CCA (find two orthogonal $d \times d$ transformation matrices for each data set to maximize their correlation), which is a linear embedding and matching procedure.

Once the mappings are learned from the training data, the learned mappings $\rho_{l}$ can be applied to match any new observations $y_{1} \in \Xi_{1}$ and $y_{2} \in \Xi_{2}$ of unknown correspondence, i.e., compute $\hat{y}_{l} = \rho_{l}(y_{l}) \in \mathbb{R}^{d}$, and declare $y_{1}$ and $y_{2}$ as matched if and only if $\hat{y}_{1}$ is sufficiently close to $\hat{y}_{2}$ in the common space. Ideally, a good matching procedure should be able to correctly identify the correspondence of the new observations, i.e., if the testing observations are truly matched in the context, the mapped points should be very close to each other in $\mathbb{R}^{d}$. If the testing observations are not matched, the mapped points should be far away from each other.

To evaluate a given matching algorithm, a natural criterion is the matching ratio used in seeded graph matching \citep{LyzinskiFishkindPriebe2014}. Assume that there exist multiple testing observations in each space; and for each testing observation $y_{1}$ in $\Xi_{1}$, there is a unique testing observation $y_{2} \in \Xi_{2}$ such that $y_{1} \sim y_{2}$. Then they are correctly matched if and only if $\hat{y}_{2}$ is the nearest neighbor of $\hat{y}_{1}$ among all other testing data from $\Xi_{2}$, and vice versa. The matching ratio equals the percentage of correct matchings, and a higher matching ratio indicates a better matching algorithm. 

The matching ratio based on nearest neighbor is often conservative, and can be a very small number when matching disparate real datasets. In practice, it is often more interesting to consider all neighbors within a small threshold, or rank multiple neighbors up to a limit. To that end, the testing power of the statistical hypothesis $H_{0}: y_{1} \sim y_{2}$ considered in \citep{PriebeMarchette2012} is another suitable criterion, which directly takes the Euclidean distance $\|\hat{y}_{1}- \hat{y}_{2}\|$ as the test statistic. To estimate the testing power for given data, we first split all observations into matched training data pairs, matched testing data pairs, and unmatched testing data pairs. After learning $\rho_{l}$ from the matched training data and applying them to all testing data, the test statistic under the null hypothesis can be estimated from the matched testing pairs, and the test statistic under the alternative hypothesis can be estimated from the unmatched testing pairs. The testing power at any type $1$ error level is directly estimated from the empirical distributions of the test statistic, and a higher testing power indicates a better manifold matching algorithm. 

We used both the testing power and the matching ratio for evaluation in the numerical experiments, and in most cases they yield the same interpretation regarding which method has a better matching performance. Note that if the critical value at a given type $1$ error level is used as a distance threshold, the testing power equals the probability that the distance between the matched pair is no larger than the distance threshold. Since the matching ratio only considers the nearest neighbor of the matched pair, the testing power is never smaller than the matching ratio.

\subsection{Main Algorithms}
\label{main1}
Our methodology is henceforth referred to as MMSJ. Algorithm~\ref{alg1} serves to learn the matching transformations from the matched training data, while algorithm~\ref{alg2} maps any testing observation onto the learned manifolds.

Given the distance matrices $\Delta_{l}$ for the training data $\{X_{l}, l=1,2\}$, we first construct an $n \times n$ binary graph $G$ by k-nearest-neighbor using the sum of normalized distance matrices $\sum_{l=1}^{2} \frac{\Delta_{l}}{\|\Delta_{l}\|_{F}}$, i.e., $G(i,j)=1$ if and only if $\sum_{l} \frac{\Delta_{l}(x_{il},x_{jl})}{\|\Delta_{l}\|_{F}}$ is among the smallest $k$ elements in the set $\{ \sum_{l} \frac{\Delta_{l}(x_{il},x_{ql})} {\|\Delta_{l}\|_{F}}, q=1,\ldots,n \}$.

Next, for each modality $X_{l}$, we calculate the shortest-path distance matrix $\Delta_{l}^{G}$ based on the normalized $\Delta_{l}$ and the joint graph $G$, i.e., solve the shortest-path problem using the weighted graph $\frac{\Delta_{l} \circ G }{\|\Delta_{l}\|_{F}}$, where $\circ$ denotes the Hadamard product. Then we apply MDS to embed $\Delta_{l}^{G}$ into $\mathbb{R}^{d}$ for each $l$, followed by the Procrustes matching to yield the matched data $\hat{X}_{l}$, i.e., the Procrustes matching finds a $d \times d$ rotation matrix by
\begin{align*}
P&=\arg\min_{P'P=I} \|P \tilde{X}_{1}-\tilde{X}_{2}\|_{F}^{2},
\end{align*}
and sets $\hat{X}_{1}=P \tilde{X}_{1}$ and $\hat{X}_{2}=\tilde{X}_{2}$, where $\tilde{X}_{l}$ denotes the embedded data by MDS.

Then each testing observation $y_{1} \in \Xi_{1}$ is mapped as follows: Given the distance between testing and training $\Delta_{1}(y_{1},X_{1})$ and the shortest-path distances for the training data $\Delta_{1}^{G}$, we first approximate the shortest-path distances $\Delta_{1}^{G}(y_{1},X_{1})$ by the respective nearest neighbors of the testing data within each modality. Then the testing data $y_{1}$ are embedded by MDS out-of-sampling (OOS) technique into $\mathbb{R}^{d}$ to yield $\tilde{y}_{1}$, followed by the Procrustes matching (i.e., $\hat{y}_{1}=P \tilde{y}_{1}$ or $\hat{y}_{2}=\tilde{y}_{2}$). Similarly for any $y_{2} \in \Xi_{2}$. Note that MMSJ merely requires the training observations from different modalities to be corresponded and of same size, but the testing observations from different modalities can be of different and arbitrary size, because they are separately mapped within each modality and of unknown correspondence. In the numerical experiments we opt to set the testing data to be of the same size for convenience of presentation and evaluation.

\begin{algorithm}
\caption{Manifold Matching using Shortest-Path Distance and Joint Neighborhood Selection (MMSJ)}
\label{alg1}
\begin{algorithmic}[1]
\Require The distance matrices $\Delta_{l}$ for the matched datasets $\{X_{l}, l=1,2\}$, the neighborhood choice $k$, and the dimension choice $d$.
\Ensure The mapped datasets $\{\hat{X}_{l} \in \mathbb{R}^{d \times n}, l=1,2$\}, the shortest-path distance $\Delta_{l}^{G}$, and the learned Procrustes transformation $P$.
\Function{MMSJ}{$\Delta_{1},\Delta_{2},k,d$}

\Linefor{$i,j:=1,\ldots,n$}{$G_{ij} \rto \sum_{l} \frac{\Delta_{l}(x_{il},x_{jl})}{\|\Delta_{l}\|_{F}}$} 
\State $G = \textsc{Rank}(G)$ \Comment{rank distances within each row}
\Linefor{$i,j:=1,\ldots,n$}{$G_{ij} \rto \mb{I}(G_{ij} \leq k)$} 
\For{$l:=1,2$}
\State $\Delta_{l}^{G} = \textsc{ShortestPath}(\frac{\Delta_{l} \circ G }{\|\Delta_{l}\|_{F}})$ 
\State $\tilde{X}_{l}=\textsc{MDS}(\Delta_{l}^{G},d)$ \Comment{embedding into $\mathbb{R}^{d}$}
\EndFor

\State $[U,S,V]=\textsc{SVD}(\tilde{X}_{2}^{T}\tilde{X}_{1})$
\State $P \rto UV^{T}$ \Comment{Procrustes matching}
\State $\hat{X}_{1}=P \tilde{X}_{1}$
\State $\hat{X}_{2}=\tilde{X}_{2}$
\EndFunction
\end{algorithmic}
\end{algorithm}

\begin{algorithm}
\caption{MMSJ on Testing Data}
\label{alg2}
\begin{algorithmic}[1]
\Require The distance vectors $\Delta_{l}(y_{l},X_{l})$ (either $l=1$ or $2$), the shortest-path distance matrices $\Delta_{l}$ and the mapped data $\hat{X}_{l}$, the learned Procrustes transformation $P$, and the neighborhood choice $k$.
\Ensure The mapped testing observation $\hat{y}_{l}$. 
\Function{MMSJ2}{$\Delta_{l}(y_{l},X_{l}),\Delta_{l}^{G},\hat{X}_{l},P,k$}
\State $G_{l} = \textsc{Rank}(\Delta_{l}(y_{l},X_{l}))$
\State $\Delta_{l}^{G}(y_{l},X_{l}) = \textsc{ShortestPath}([\Delta_{l}^{G} | \Delta_{l}(y_{l},X_{l}) \circ G_{l}])$ 
\State $\tilde{y}_{l}=\textsc{MDS-OOS}(\hat{X}_{l},\Delta_{l}^{G}(y_{l},X_{l}))$
\If{$l=1$}
\State $\hat{y}_{1}=P \tilde{y}_{1}$
\Else
\State $\hat{y}_{2}=\tilde{y}_{2}$
\EndIf
\EndFunction
\end{algorithmic}
\end{algorithm}

\begin{figure}[ht]
\centering
\begin{tikzpicture}[node distance=1.5cm]
\node (in1) [node, xshift=4cm] {$[X_{1}] \in \Xi_{1}$};
\node (in2) [node, xshift=8cm] {$[X_{2}] \in \Xi_{2}$};
\node (D1) [node, below of=in1] {$\Delta_{1}$};
\node (D2) [node, below of=in2] {$\Delta_{2}$};

\node (D12) [node, below of=D1,xshift=2cm] {$\frac{\Delta_{1}}{\|\Delta_{1}\|_{F}}+\frac{\Delta_{2}}{\|\Delta_{2}\|_{F}}$};
\node (G) [node, below of=D12] {$G$};
\node (DG1) [node, left of=G,xshift=-0.5cm] {$\Delta_{1}^{G}$};
\node (DG2) [node, right of=G,xshift=0.5cm] {$\Delta_{2}^{G}$};
\node (X1) [node, below of=DG1] {$\tilde{X}_{1} \in \mathbb{R}^{d \times n}$};
\node (X2) [node, below of=DG2] {$\tilde{X}_{2} \in \mathbb{R}^{d \times n}$};
\node (end) [node, below of=X1,xshift=2cm] {$\hat{X}_{l} \in \mathbb{R}^{d \times n}$};

\draw [arrow] (in1) -- (D1);
\draw [arrow] (in2) -- (D2);
\draw [arrow] (D1) -- (D12);
\draw [arrow] (D2) -- (D12);
\draw [arrow, align=right] (D12) -- node[anchor=west] {\small Joint Graph}(G);
\draw [arrow, align=right] (D2) -- node[anchor=west] {\small Shortest-Path}(DG2);
\draw [arrow] (D1) -- (DG1);
\draw [arrow] (G) -- (DG1);
\draw [arrow] (G) -- (DG2);
\draw [arrow, align=right] (DG2) -- node[anchor=west] {\small Embedding}(X2);
\draw [arrow] (DG1) -- (X1);
\draw [arrow] (X2) -- node[anchor=west] {\small Matching}(end);
\draw [arrow] (X1) -- (end);
\end{tikzpicture}
\caption{Flowchart for Algorithm~\ref{alg1}}
\label{flow}
\end{figure}
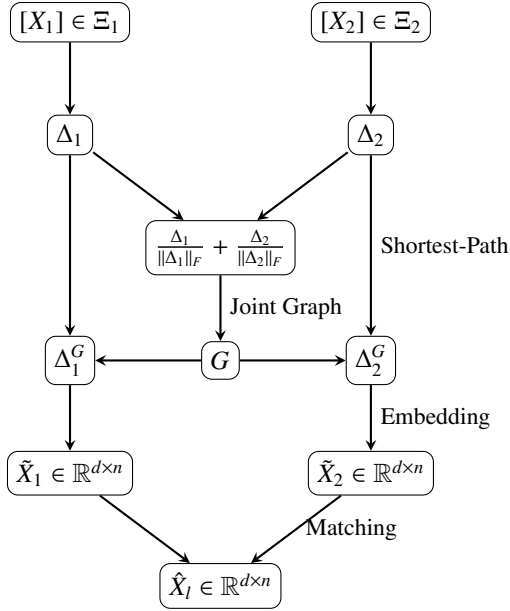

To better visualize the process, we summarize the main algorithm in the flowchart of Figure~\ref{flow}. 

\subsection{Implementation Details}
\label{discuss}
In this subsection, we elaborate on various implementation details of MMSJ.

MMSJ starts with two distance matrices rather than the sample observations directly, which means it is directly applicable to multiple modalities of different feature dimensions, as long as a distance metric can be defined for each modality. Although there is no limitation on applying the algorithm once the distances are given, the actual matching performance is clearly dependent on the choice of the metric. The most common choice is the Euclidean distance, or $L^{p}$ metrics in general. Other similarity or dissimilarity measures may be more appropriate in certain domains, such as the cosine distance for text data (see Section~\ref{wikiReal}), or suitable kernels for structured data \citep{HofmannEtAl2008}. 

The joint neighborhood graph ensures consistent neighborhood selection when at least one of the modalities are nonlinear or noisy, and is intuitively better than two separate neighborhood graphs for matching. Alternatively, one may use a weighted sum of distances or a rank-based method to derive the joint neighborhood graph instead. Note that joint neighborhood requires the distance matrices of the training data to be properly scaled in advance, but is clearly not applicable to the testing data with unknown correspondence.

Using the joint neighborhood graph allows the resulting shortest-path distance to utilize the dependency structure of the training data. Computationally, the shortest-path distance matrix can be effectively implemented by Floyd's algorithm or Dijkstra's algorithm \citep{TenenbaumSilvaLangford2000}, which can be further sped up by choosing a small set of landmark points \citep{SilvaTenenbaum2003, BengioEtal2003}. Theoretically, the shortest-path distance can recover the geodesic distance of isometric manifolds with high probability under certain sampling conditions \citep{BernsteinEtAl2000, SilvaTenenbaum2003}. When embedding the testing data, we essentially treat the training data as landmark points and only compute the shortest-path distances from the testing data to the training data.

Embedding the shortest-path distances into the Euclidean space followed by matching is a standard procedure. Alternatively, one may match the embeddings by CCA or joint MDS, as discussed in \citep{PriebeMarchette2012, FishkindShenPriebe2016}. The advantages of MMSJ mostly lie in joint neighborhood and shortest-path distance; in fact, MMSJ always exhibits significant improvement, no matter which matching method to use. Thus we mainly consider the Procrustes matching for ease of presentation in the paper. For the testing data, they are embedded by out-of-sample MDS, which is a standard technique for MDS and kernel PCA \citep{ScholkopfSmolaMuller1998,BengioEtal2003,TrossetPriebe2008} and more efficient than re-embedding all training and testing data. After all testing data are mapped onto the manifolds by the learned Procrustes matching, we may evaluate the matching performance as described in Section~\ref{bg}.

In terms of computation speed, suppose $n$ is the sample size of training pairs, and $n'$ is the size of all testing data. The running time complexity of MMSJ is $O(n^2+ n n')$, assuming the distance matrices are already given and the shortest-path step uses the fast landmark approximation. The only overhead is the distance matrix construction, which takes an additional $O(n^2 d+ nn' d)$, where $d$ denotes the maximal feature dimension among all modalities. Therefore MMSJ is computationally efficient for high-dimensional data or data with large amount of testing observations.

To compare with MMSJ, we use the standard procedure that embed each modality separately by MDS / Isomap / LLE / LTSA, followed by Procrustes matching. Note that MDS / Isomap / LLE can all operate directly on a distance matrix, but some nonlinear algorithms like LTSA have to start with the Euclidean data rather than a distance measure. Thus, if only the distance matrices are available, the distance matrices are pre-embedded into a Euclidean space $\mathbb{R}^{d'}$ with $d' \geq d$ by MDS, followed by embedding into $\mathbb{R}^{d}$ via LTSA and matching by Procrustes.

\subsection{Discussions}
\label{discuss2}

In this section, we offer further discussions on factors that can affect the matching performance.

In general, other than which matching method to apply, the matching performance is further dependent on how match-able the actual data sets are. If the observations from different modalities are strongly linearly correlated (a simple example is two almost identical databases with trivial manual errors), the matching performance will likely be perfect for any matching method; if the two modalities are related via either a very complex or non-existent transformation (such that they appear to be almost independent, e.g., weather and stock price on the same day), there is no hope to recover any meaningful matching regardless of what method to use. However, if there are certain local information that are shared by different modalities (e.g., the Swiss roll example in Section~\ref{numer}), that is the situation that MMSJ may improve over other methods by extracting useful local correspondence via the joint graph and shortest-path distance. 

Assuming the given modalities can be reasonably combined for matching, the size and selection of the training observations are critical for all matching methods (or generally any supervised learning task): If the given training data are representative of the underlying manifold, then it is like to identify the correct matching for any number of testing observations; but if the training data only reveal part of geometry, the matching performance will take a hit as testing observations increase. Of course, given the training data, whether the underlying manifold can be correctly recovered is also dependent on the embedding method. So alternatively, the advantage of MMSJ can be viewed as requiring much less training data to capture the manifolds for matching, e.g., Figure~\ref{fig2}(A) can be interpreted as that MMSJ only takes around $1000$ observations to uncover the intrinsic manifolds for perfect matching, while all other benchmarks require much larger sample size.

Another performance factor is the out-of-sampling technique. It has been popularly used as a fast approximation for distance-based embedding, and often called the Nystrom approximation \cite{Mahoney2005} in numerical analysis. Clearly, different sampled landmark points will yield different embeddings of the out-of-sample observations, so the matching performance can be affected if the landmark points are not chosen appropriately. It turns out that the role of landmark points is similar to the role of the training data: if the in-sample landmark data approximately represent the whole manifold, then the out-of-sample embedded points are generally faithful to the original geometry; but if the in-sample data fail to reflect the manifold, then the out-of-sample embedded observations will have a larger discrepancy. Therefore, in order to minimize the effect of OOS in the matching task, the suggested strategy is to take all training data as the in-sample observations and embed the testing data via OOS, which is applied to MMSJ and all benchmarks in the experiments.

\section{Numerical Experiments}
\label{numer}
In this section, we demonstrate the numerical advantages of MMSJ. Overall, we observed that our method is significantly better than all the benchmarks (MDS, Isomap, LLE, and LTSA) in matching ratio and testing power under various simulated settings and real experiments.

\subsection{Swiss Roll Simulation}
\begin{figure}[htbp]
\centering
\includegraphics[width=0.5\textwidth]{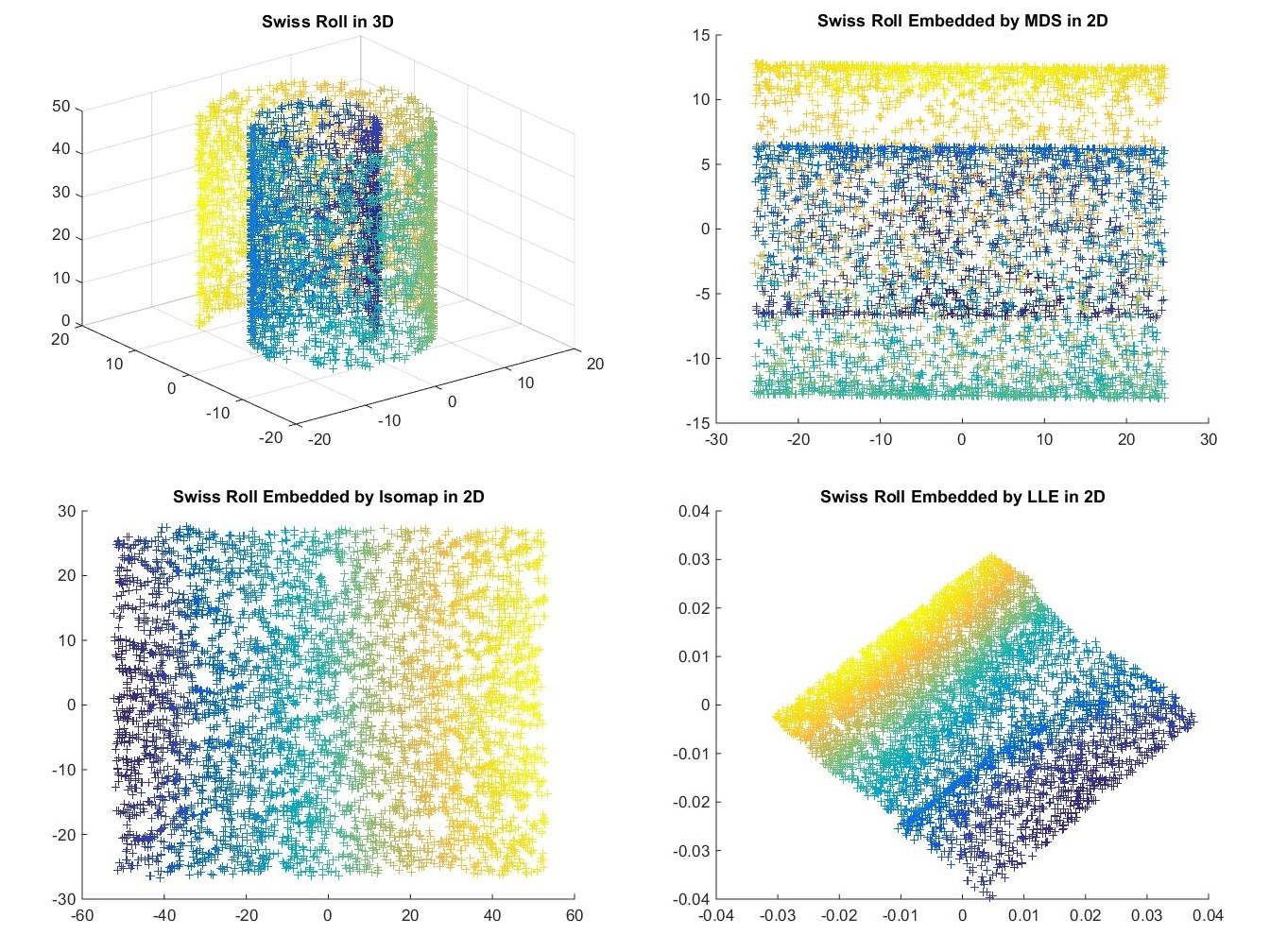}
\caption{The 3D Swiss roll dataset (top left), its 2D embedded data by MDS (top right), 2D embedding by Isomap at neighborhood size $k=10$ (bottom left), and 2D embedding by LLE at $k=10$ (bottom right).}
\label{fig1}
\end{figure}

The Swiss roll data from \citep{TenenbaumSilvaLangford2000} is a 3D dataset representing a nonlinear manifold, which is intrinsically generated by points on a 2D linear manifold. Figure~\ref{fig1} shows the 3D Swiss roll data with $5000$ points in colors, along with its 2D embeddings by MDS, Isomap, and LLE. Clearly, MDS fails to recognize the nonlinear geometry while both Isomap and LLE succeed. However, the LLE embedding has a distorted geometry, while the Isomap embedding is more faithful to the underlying 2D linear manifold.

For the first simulation, we match the 3D Swiss roll with its underlying 2D linear manifold at varying training data size. A total of $n$ points from the 3D Swiss roll are randomly generated to construct the first modality $X_{1}$, and the corresponding points on the underlying 2D linear manifold are taken as the second modality $X_{2}$. Thus $X_{1}$ and $X_{2}$ are matched training data with distinct geometries. Once the training data are matched, we embed and apply the learned mappings to new testing observations $y_{1}$ and $y_{2}$ in each space.

\begin{figure*}
  \centering
  \begin{tabular}{@{}p{\linewidth}@{\quad}p{\linewidth}@{}}
	\centering
    \subfigimg[width=0.32\linewidth]{A}{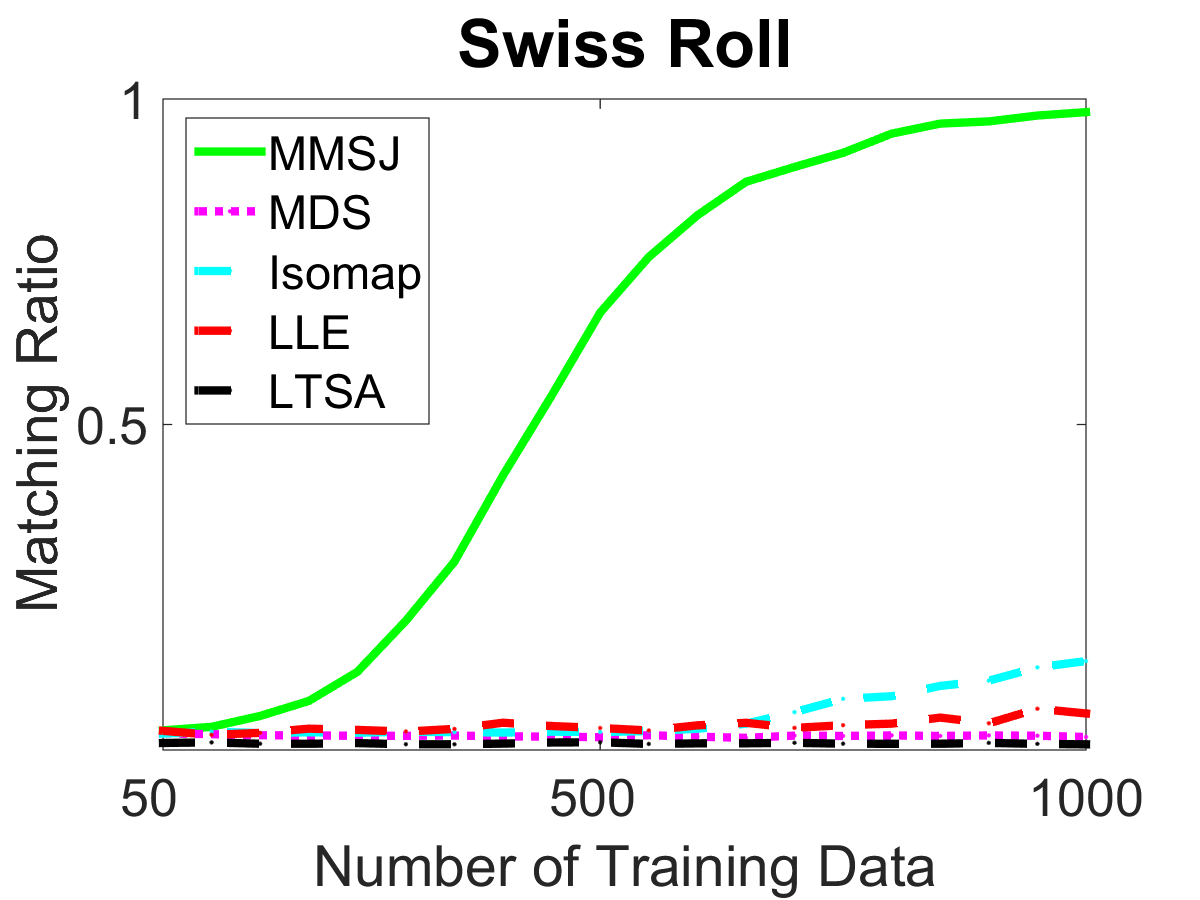}
    \subfigimg[width=0.32\linewidth]{B}{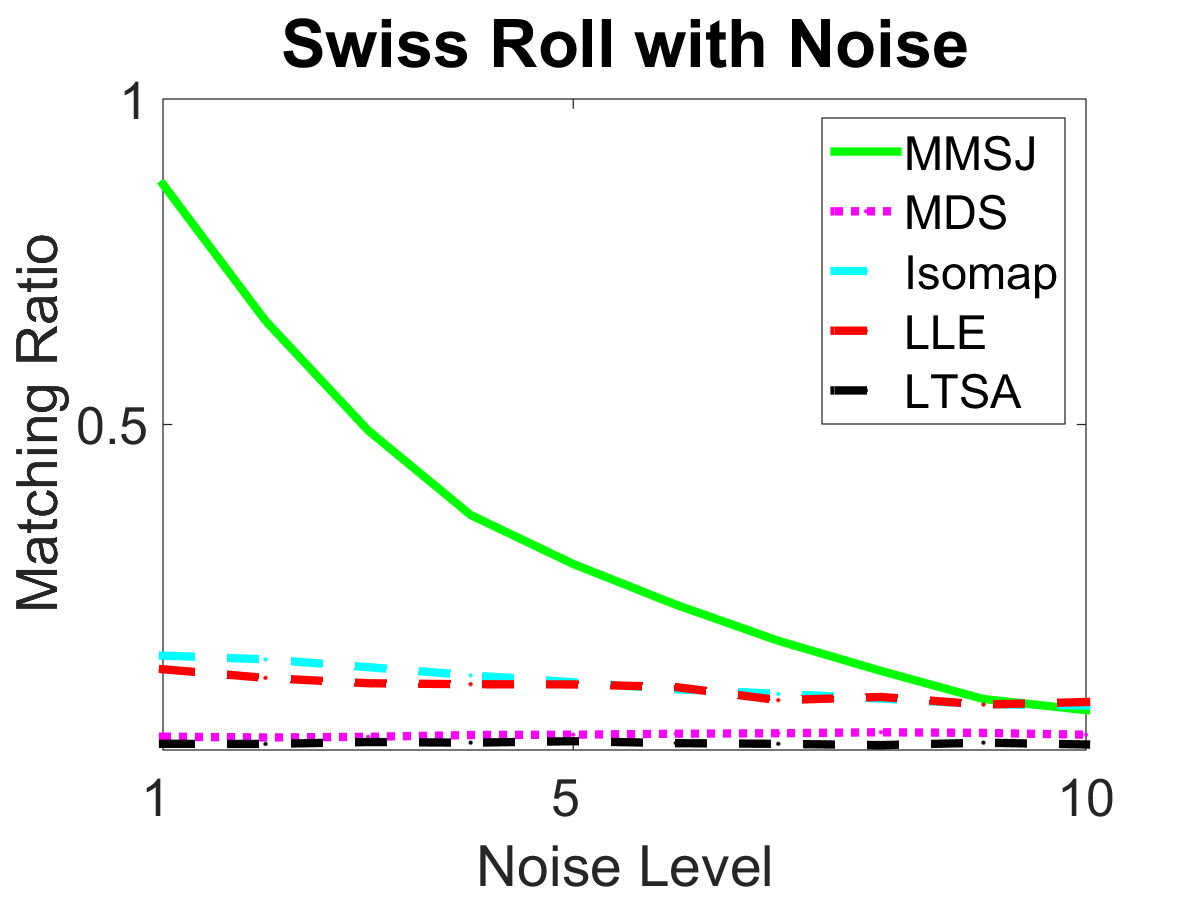}
    \subfigimg[width=0.32\linewidth]{C}{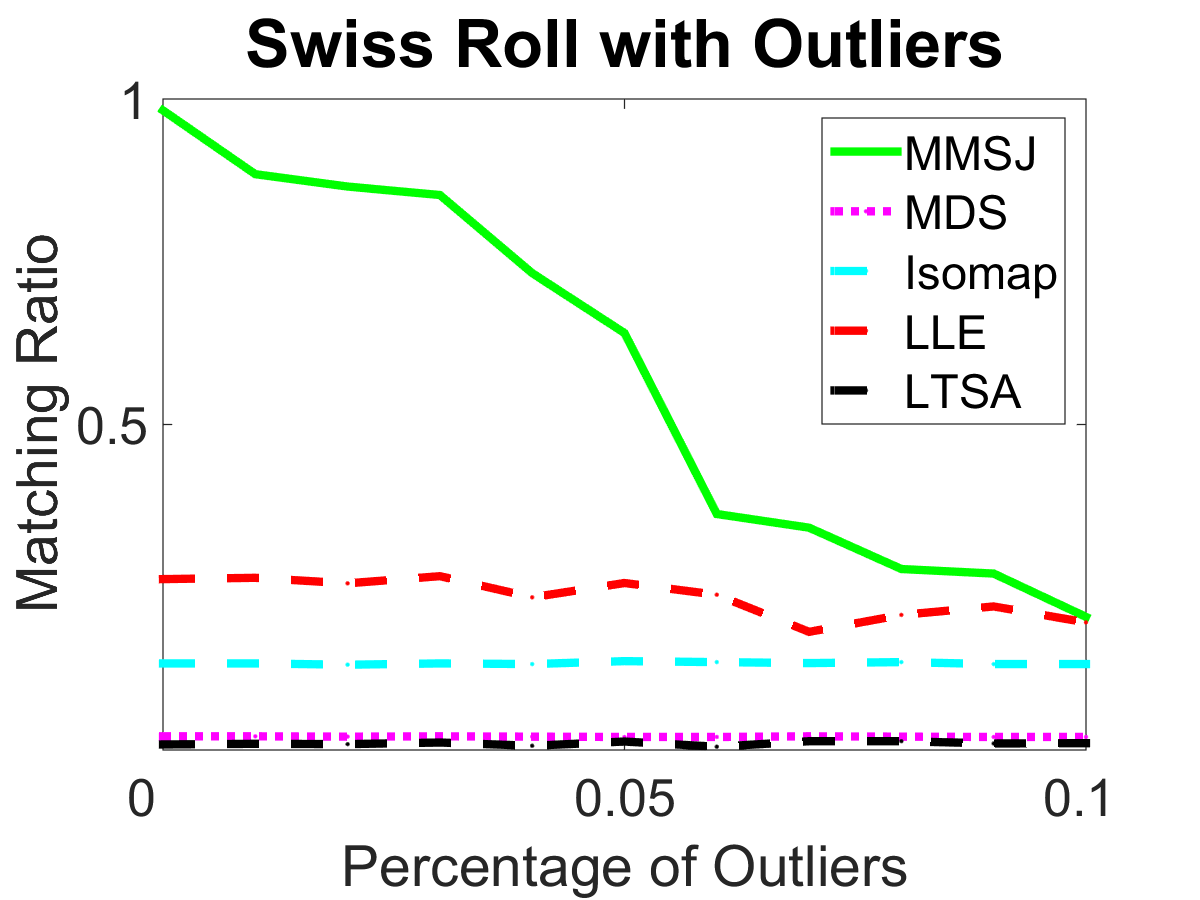}
  \end{tabular}
  \caption{ Matching Ratio of 3D Swiss Roll versus its 2D Underlying Linear Manifold.
(A) Matching Ratio with respect to Increasing Size of Training Data.
(B) Matching Ratio with respect to Increasing Noise at $n=1000$.
(C) Matching Ratio with respect to Growing Number of Outliers at $n=1000$. }
\label{fig2}
\end{figure*}

We set the neighborhood size as $k=10$, the dimension choice as $d=2$, and generate $n'=100$ testing pairs to compute the matching ratio. We repeat the above process for $100$ Monte-Carlo replicates, and show the average matching ratio in Figure~\ref{fig2}(A) with respect to increasing training data at $n=50,100,\ldots,1000$. 
The MMSJ algorithm exhibits a significant advantage over all other algorithms: it enjoys a better matching ratio from small sample size onwards, and achieves almost perfect matching as sample size grows.

Next, we check the robustness of the manifold matching algorithms against noise, by adding white noise to the linear modality $X_{2}$ and fixing the training data size to $n=1000$. The noise is independently and identically distributed as $Normal(0, \epsilon I_{2 \times 2})$, and the same testing procedure is applied to compute the matching ratio at increasing noise levels. The results are plotted in Figure~\ref{fig2}(B) with respect to $\epsilon = 0,1,2,\ldots,10$. The MMSJ algorithm is clearly better than all the benchmarks as the noise level increases.

For the third simulation, we consider an outlier scenario that randomly permutes a portion of the training data. For $\epsilon=0,0.01,\ldots,0.1$, we randomly pick $\epsilon n$ training data from $X_{2}$ and permute their indices, such that those training data are no longer matched with the corresponding observations from $X_{1}$, i.e., there exists $\epsilon n$ outliers in the training data. Fixing $n=1000$, we apply the same testing procedure and plot the matching ratio in Figure~\ref{fig2}(C) with respect to the outlier percentage $\epsilon$. The MMSJ algorithm again has better matching ratio throughout increasing $\epsilon$; and all methods exhibit insignificant matching ratio when the outlier percentage $\epsilon$ reaches beyond $0.1$, implying that the matching task may benefit significantly from excluding outliers prior to matching.

\subsection{Wikipedia Articles Experiments}
\label{wikiReal}
In this section, we apply the manifold matching algorithm to match disparate features of Wikipedia articles. The raw data contains $1382$ pairs of articles from Wikipedia English and the corresponding French translations, within the 2-neighborhood of the English article ``Algebraic Geometry". On Wikipedia, the same articles of different languages are almost never the exact translations of each other, because they are very likely written by different people and their contents may differ in many ways.

For the English articles and their French translations, a text feature and a graph feature are collected separately under each language. For the texts of each article, we use latent semantic indexing (LSI) (i.e., first construct a term-document matrix to describe the occurrences of terms in documents, then apply the low-rank approximation to the term matrix to $100$ dimensions by singular value decomposition, see \citep{DeerwesterDumais1990} for details) followed by cosine dissimilarity to construct two dissimilarity matrices $TE$ and $TF$ (representing the English texts and French texts). For the networks, two shortest-path distance matrices $GE$ and $GF$ (representing the English graph and French graph) are calculated based on the Internet hyperlinks of the articles under each language setting, with any path distance larger than $4$ imputed to be $6$ to avoid infinite distances and scaling issues.

Therefore, there exist four different modalities for pairs of Wikipedia articles on the same topic, making $TE$, $TF$, $GE$, and $GF$ matched in the context. Furthermore, as the text matrices are derived by cosine similarity while the graph matrices are based on the shortest-path distance with imputation, the former probably have nonlinear geometries while the latter are linear from the view of our matching algorithm.

For each Monte-Carlo replicate, we randomly pick $n=500$ pairs of training observations, $100$ pairs of testing matched observations, and $100$ pairs of testing unmatched observations for evaluation. The parameters are set as $k=20$, $d=10$, $d'=50$ (for LTSA only), and the manifold matching algorithms are applied for every possible combination of matching two modalities. We perform a total of $100$ Monte-Carlo replicates. The mean matching ratio is reported in Table~\ref{table:wikiAcc}, and the estimated testing power is presented in Table~\ref{table:wikiPower} at type $1$ error level $0.05$.

Clearly, MMSJ achieves the best performance throughout all combinations. From the tables and figures, we further observe that without using shortest-path distance or joint neighborhood, separate nonlinear embeddings from LLE or LTSA are worse than the linear MDS embeddings in matching. Isomap does fairly well in the testing power, as it also uses shortest-path distance, but it can still be occasionally similar or slightly inferior to MDS in the matching ratio. Our proposed MMSJ algorithm is consistently the best manifold matching algorithm in both the testing power and the matching ratio.

\begin{figure*}
  \centering
  \begin{tabular}{@{}p{\linewidth}@{\quad}p{\linewidth}@{}}
	\centering
    \subfigimg[width=0.32\linewidth]{A}{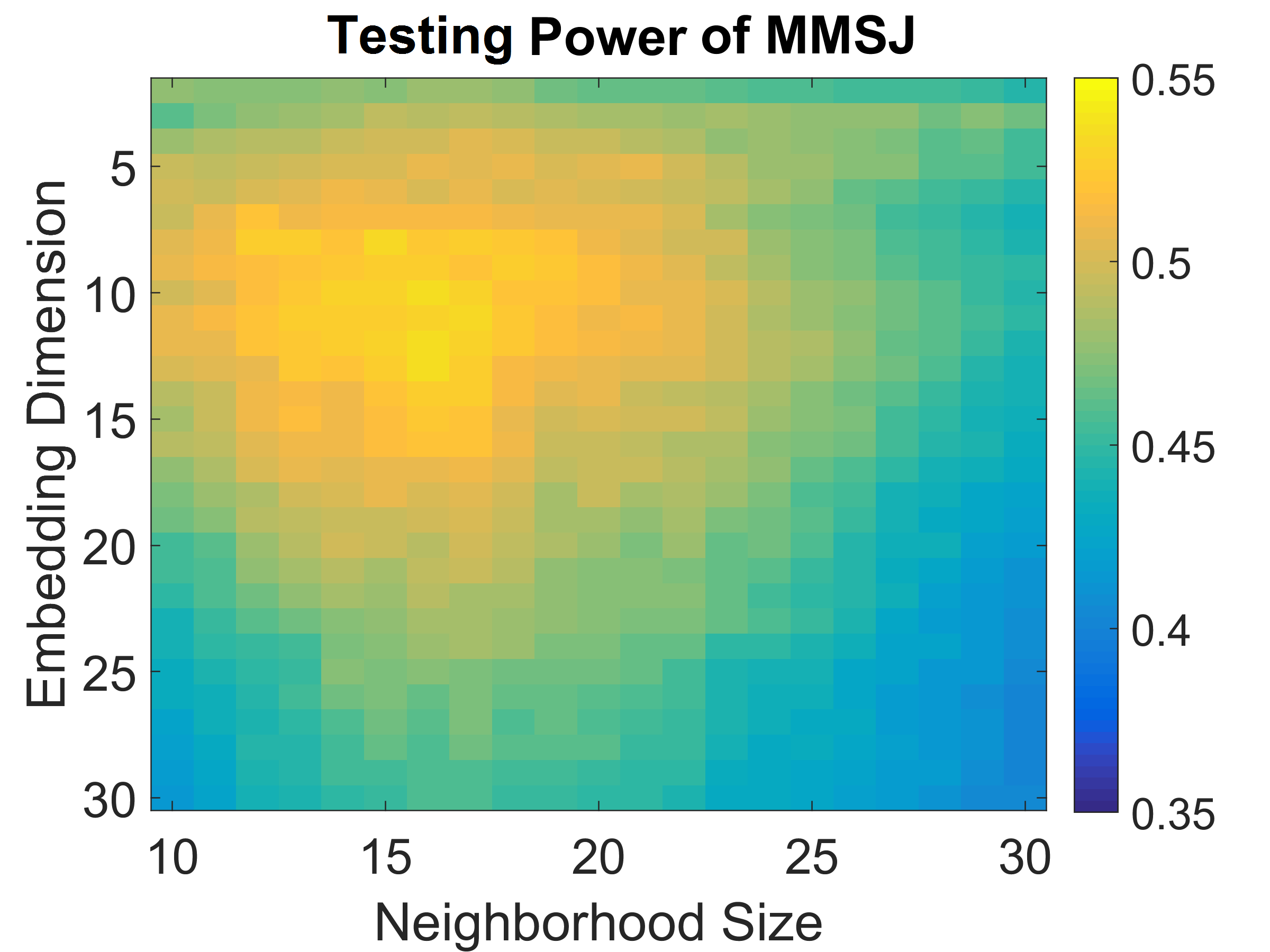}
    \subfigimg[width=0.32\linewidth]{B}{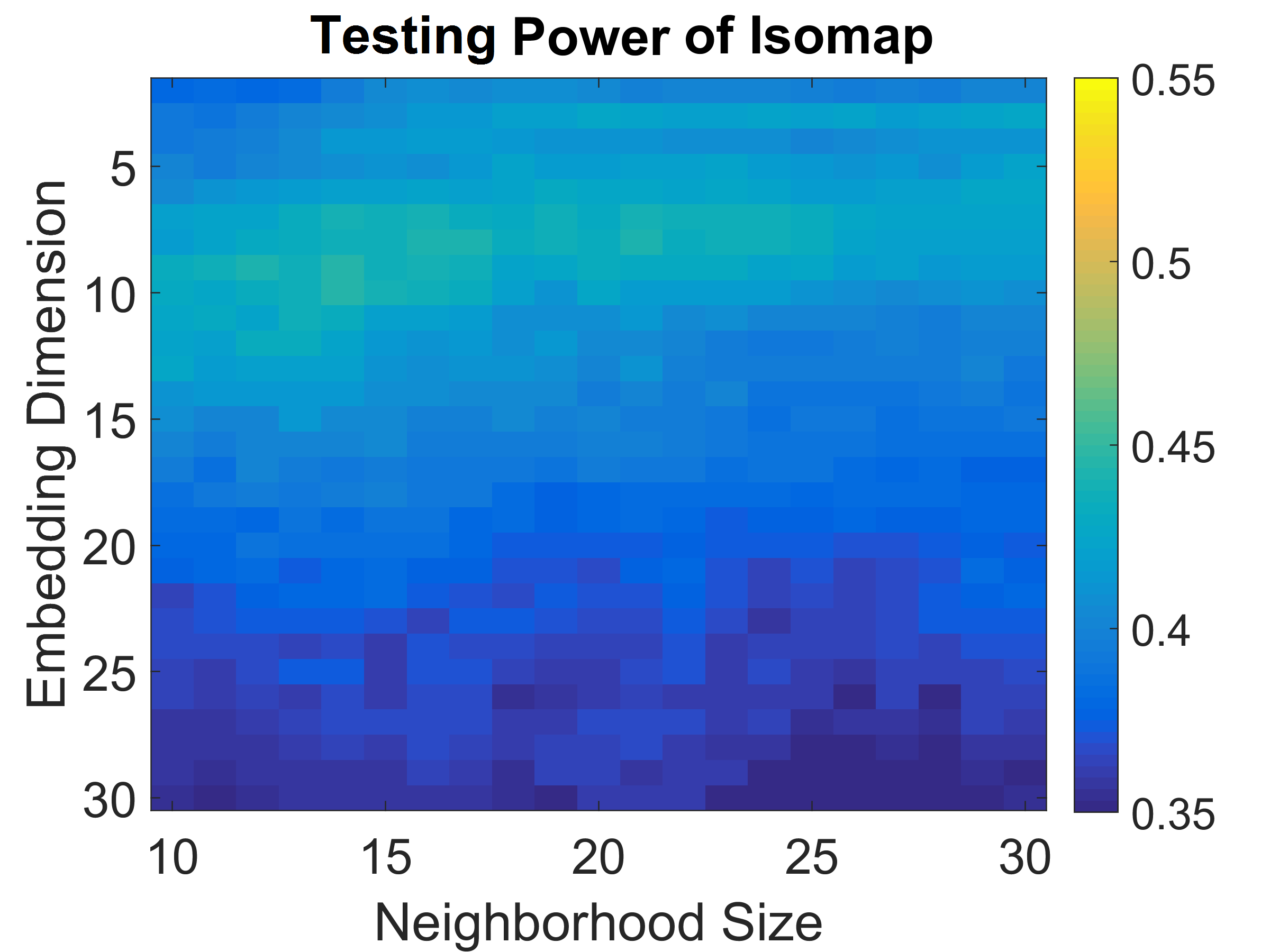}
    \subfigimg[width=0.32\linewidth]{C}{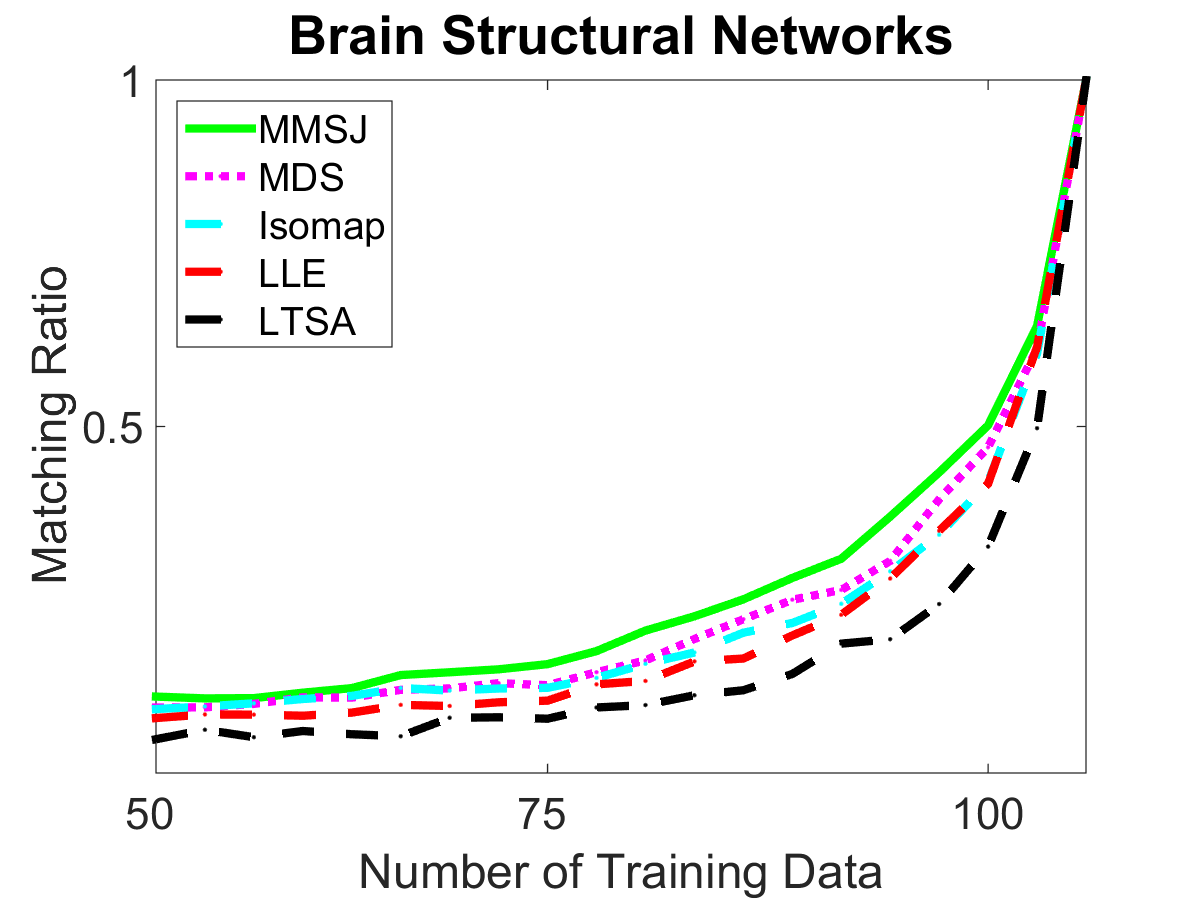}
  \end{tabular}
  \caption{ (A) The Testing Power of MMSJ for Matching Wikipedia English Text and English Graph with respect to Different Dimension Choices and Neighborhood Sizes at Type $1$ Error Level 0.05.
(B) Same as (A) but for The Testing Power of Isomap.
(C) Matching Ratio of Brain Structural Networks with respect to Increasing Size of Training Data.}
\label{figRealSurf}
\end{figure*}

For the next experiment, we demonstrate that MMSJ algorithm is also robust against misspecification of parameters. The first two panels of Figure~\ref{figRealSurf} plot the MMSJ and Isomap testing powers (the best two algorithms in our matching experiments) for matching $(TE,GE)$, against different choices of $d \in [2,30]$ and $k \in [10,30]$. Comparing the two panels not only shows that MMSJ is always better than Isomap in matching, but it also attains comparable testing power in a wide range of parameter choices.
The same robustness holds for MMSJ under all other matching combinations. 

\begin{Table}
\centering
\captionof{table}{Wikipedia Documents Matching Ratio}
\label{table:wikiAcc}%
\begin{tabular}{|c||c|c|c|c|c|}
\hline
Modalities & MMSJ & MDS & Isomap & LLE & LTSA \\
\hline
$(TE, TF)$ & $\textbf{0.2942}$  & $0.2546$ & $0.2003$ & $0.1265$ & $0.0491$\\
\hline
$(TE, GE)$ & $\textbf{0.1209}$  & $0.0675$ & $0.0866$ & $0.0143$ & $0.0260$\\
\hline
$(TF, GF)$ & $\textbf{0.0624}$  & $0.0419$ & $0.0522$ & $0.0134$ & $0.0144$\\
\hline
$(GE, GF)$ & $\textbf{0.1347}$  & $0.1280$ & $0.1081$ & $0.0157$ & $0.0236$\\
\hline
$(TE, GF)$ & $\textbf{0.0677}$  & $0.0429$ & $0.0560$ & $0.0132$ & $0.0138$\\
\hline
$(TF, GE)$ & $\textbf{0.0946}$  & $0.0545$ & $0.0698$ & $0.0132$ & $0.0238$\\
\hline
\end{tabular}
\end{Table}

\begin{Table}
\centering
\captionof{table}{Wikipedia Documents Testing Power at Type $1$ Error Level 0.05}
\label{table:wikiPower}
\begin{tabular}{|c||c|c|c|c|c|}
\hline
Modalities & MMSJ & MDS & Isomap & LLE & LTSA \\
\hline
$(TE, TF)$ & $\textbf{0.8124}$  & $0.4974$ & $0.7476$ & $0.3594$ & $0.1930$\\
\hline
$(TE, GE)$ & $\textbf{0.5184}$  & $0.2563$ & $0.4255$ & $0.0948$ & $0.1116$\\
\hline
$(TF, GF)$ & $\textbf{0.2782}$  & $0.1128$ & $0.1877$ & $0.0903$ & $0.1028$\\
\hline
$(GE, GF)$ & $\textbf{0.3108}$  & $0.2141$ & $0.2485$ & $0.0961$ & $0.1063$\\
\hline
$(TE, GF)$ & $\textbf{0.3199}$  & $0.1130$ & $0.2141$ & $0.0923$ & $0.1021$\\
\hline
$(TF, GE)$ & $\textbf{0.4464}$  & $0.2114$ & $0.3595$ & $0.0943$ & $0.1064$\\
\hline
\end{tabular}
\end{Table}

\subsection{Brain Structural Networks}
In this section, we assess the matching performance via brain structural networks.  There are a total of  $n=109$ subjects, each with diffusion weighted magnetic resonance imaging (MRI) data. For the raw brain imaging data, we derived two different modalities.  For each scan, (i) process diffusion and structural MRI data via MIGRAIN, a pipeline for estimating brain networks from diffusion data \cite{GrayRoncal2013}, (ii) compute the distance between brain networks using the semi-parametric graph test statistic \cite{Sussman2013,Tang2016}, then embed each graph into two dimensions and align the embeddings via a Procrustes analysis. The Euclidean distance is used on both modalities. 

Therefore the first modality seems like a more faithful representation of the brain imaging, while the second modality is inherently a graph representation of the brain structure. Although these two modalities are merely different transformations of the same raw data, they are distinct in many aspects such that there is no guarantee that one can recover their underlying geometry or succeed the matching task via machine learning algorithms.

For each Monte-Carlo replicate, we randomly pick $n$ pairs of matched observations for training, with all remaining sample observations for testing. The parameters are set as $k=7$, $d=2$, $d'=2$, with a total of $100$ Monte-Carlo replicates. The mean matching ratio is shown in Figure~\ref{figRealSurf}(C) with respect to increasing size of training data. It is clear that MMSJ is the best matching method among all algorithms, and all matching ratios improve significantly as the training data size increases relative to the testing data size.

\section{Concluding Remarks}
\label{conclu}
In summary, we propose a nonlinear manifold matching algorithm using shortest-path distance and joint neighborhood selection. The algorithm is straightforward to implement, efficient to run, and achieves superior matching performance. It is able to significantly improve the testing power and matching ratio for multiple modalities of distinct geometries, and is robust against noise, outliers, and model misspecification. Our experiments indicate that the shortest-path distance and joint neighborhood selection are two key catalysts behind the improvement of the matching performance.

There are a number of potential extensions of this work. First, pursuing theoretical aspects of the manifold matching task is a very challenging but rewarding task: so far there is a very limited number of literatures even for manifold learning of a single modality, and no nonlinear transformation can always recover the linear manifold under a wide range of geometries. On the other hand, the task of matching multiple modalities is unique on its own. As a first step towards better theoretical understanding, we successfully proved in \cite{ShenEtAl2017} that testing dependence via local correlations (which makes use of joint neighborhood graph and local distance in a similar manner) can successfully detect almost all relationships as sample size grows large, which shall shed more lights into the consistency of the matching task and may further advance the MMSJ algorithm.

Second, MMSJ requires a pair of metrics (or distances) for each modality. In this work we assume such metrics are pre-defined by domain knowledge, or use the Euclidean distance otherwise. If an optimal metric can be reasonably selected for each modality, it is likely to further boost the performance of MMSJ. From another point of view, if we are given high-dimensional or structured data (say graphs or images) to match, the algorithm may further benefit from first carrying out an appropriate feature selection down to certain landmark features \citep{JJ2008, ConteEtAl2004, FioriEtAl2013}, then use MMSJ. This quest is a valuable future direction to work on.

\section*{Acknowledgment}
\addcontentsline{toc}{section}{Acknowledgment}
This work is partially supported by the National Security Science and Engineering Faculty Fellowship (NSSEFF),
 the Johns Hopkins University Human Language Technology Center of Excellence (JHU HLT COE), and the
 XDATA program of the Defense Advanced Research Projects Agency (DARPA) administered through Air Force Research Laboratory contract FA8750-12-2-0303. This work is also supported by the Defense Advanced Research Projects Agency (DARPA) SIMPLEX program through SPAWAR contract N66001-15-C-4041 and DARPA GRAPHS N66001-14-1-4028.

The authors would like to thank the reviewers for their insightful and valuable suggestions in improving the exposition of the paper.

\bibliographystyle{ieeetr}
\bibliography{references}



\end{document}